\documentclass{article}
\usepackage{spconf,amsmath,graphicx, bm}
\usepackage{amssymb}
\usepackage{enumitem}
\usepackage{multirow}
\usepackage{float}
\usepackage{graphicx}
\usepackage{stmaryrd}
\usepackage{mathabx}
\usepackage[dvipsnames,table,xcdraw]{xcolor}
\usepackage[pagebackref=true,breaklinks=true,colorlinks=true,bookmarks=false, linkcolor=Maroon, citecolor=PineGreen]{hyperref}
\usepackage{url}            % simple URL typesetting
\usepackage{bibspacing}
\title{Progressive Multi-stage Feature Mix for Person Re-Identification}
\name{Yan Zhang$^1$, Binyu He$^1$, Li Sun$^{1,2}$\sthanks{Corresponding author (Email: sunli@ee.ecnu.edu.cn). This work is supported by the the Science and Technology Commission of Shanghai Municipality (No.19511120800), and the Open Project Program of the State Key Laboratory of Mathematical Engineering and Advanced Computing.}, and Qingli Li$^1$}
\address{$^1$Shanghai Key Laboratory of Multidimensional Information Processing, %\\$^2$Key Laboratory of Advanced Theory and Application in Statistics and Data Science, 
\\East China Normal University, 200241 Shanghai, China
\\$^2$State Key Laboratory of Mathematical Engineering and Advanced Computing, 214125 Wuxi, China}
\begin{document}
%\ninept
%
\maketitle
% -----------------------------------------------------------------------
\begin{abstract}
    Image features from a small local region often give strong evidence in person re-identification task. However, CNN suffers from paying too much attention on the most salient local areas, thus ignoring other discriminative clues, \emph{e.g.}, hair, shoes or logos on clothes. %BDB proposes to randomly drop one block in a batch to enlarge the high response areas. Although BDB has achieved remarkable results, there still room for improvement. 
    In this work, we propose a \emph{Progressive Multi-stage feature Mix network} % Progressive Classifier-guided feature Mix network (PCM)
    (PMM), which enables the model to find out the more precise and diverse features in a progressive manner. Specifically, 
    \emph{(\romannumeral1)} to enforce the model to look for different clues in the image, we %propose to remove the most salient features stage by stage, 
    adopt a multi-stage classifier %and each stage is trained on the same task, 
    and expect that the model is able to focus on a complementary region in each stage. % thus we can integrate diverse features from different stages to form the final image representation. 
    \emph{(\romannumeral2)} %BDB expands the high response area, but also introduces redundant information. 
    %Inspired by Attentive CutMix, 
    we propose an Attentive feature Hard-Mix (A-Hard-Mix) to %suppress and disturb the salient features
    replace the salient feature blocks by the negative example in the current batch, whose label is different from the current sample.
    \emph{(\romannumeral3)} extensive experiments have been carried out on reID datasets such as the Market-1501, DukeMTMC-reID and CUHK03, showing that the proposed method can boost the re-identification performance significantly. Project code: \url{https://github.com/crazydemo/Progressive-Multi-stage-Feature-Mix-for-\\Person-Re-Identification}

\end{abstract}
\section{Introduction}
      
    \begin{figure}[th]
		\centering
		\includegraphics[width=0.5\textwidth]{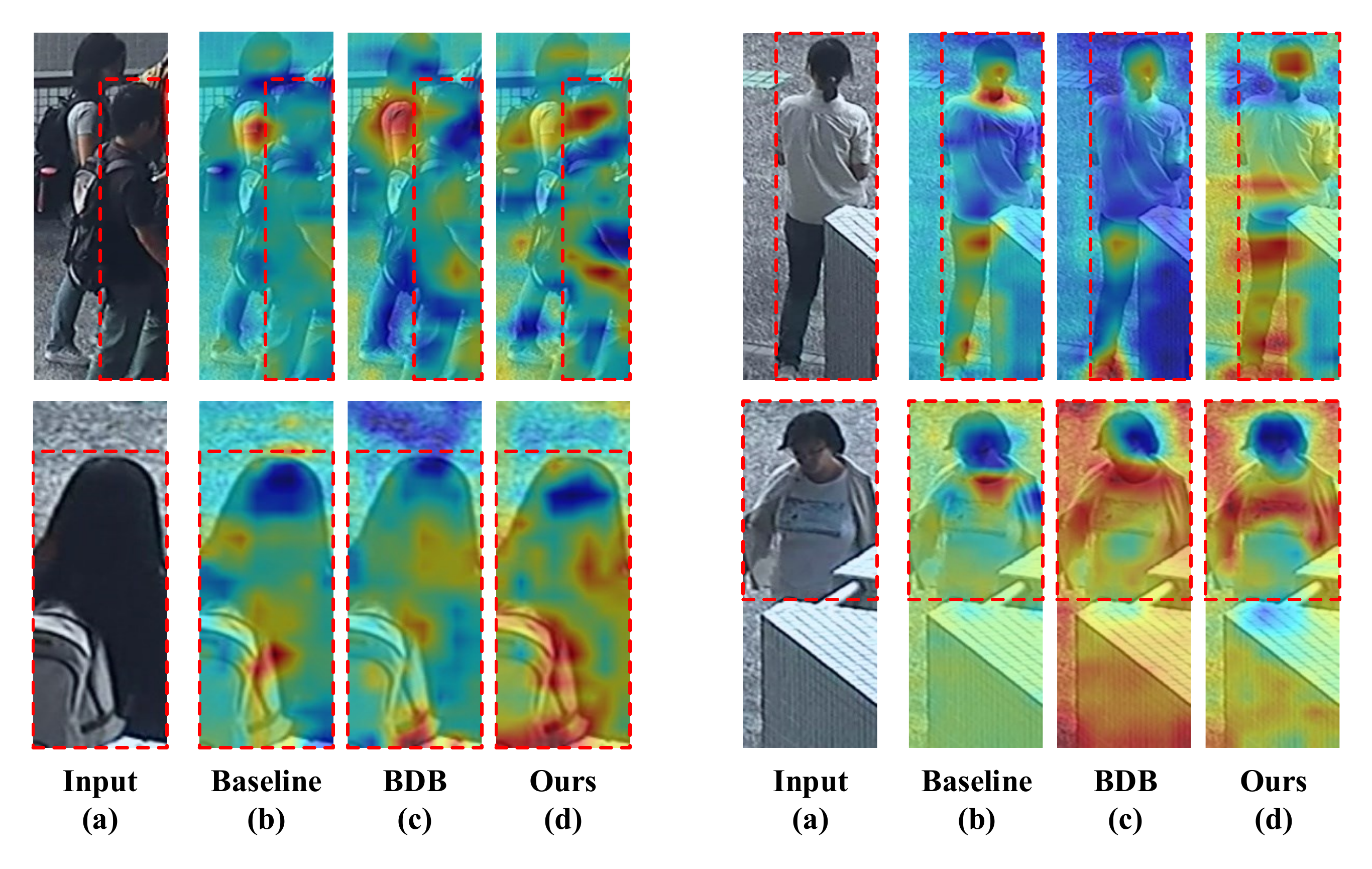}
		\caption{The visualizations of the gradient based class activation map of baseline (Resnet50+GAP), BDB \cite{BFE} and our approach. Red boxes indicate the target attention area.}
		\label{fig:impact results}
	\end{figure}
	
	Person re-identification (reID) has various potential applications in the video surveillance and security system. It aims to identify the same person from multiple detected pedestrian images, typically captured by different non-overlapping cameras. ReID is challenging mainly due to %In order to tackle with
	the partial occlusions, viewpoint changes, and illumination variations. Therefore, most of the recent works focus on %the sophisticated part feature learning \cite{topology, su2017pose}, 
	learning reliable features from local parts \cite{topology, su2017pose, kumar2017pose, zheng2019pose, kalayeh2018human},  with the extra pose or segmentation labels to help mine the clues invariant to these factors. BDB \cite{BFE} proposes a simple random block drop scheme with a multi-branch classifier to learn more robust and general feature representations, without requiring the extra labels. %However, 
	As shown in Figure \ref{fig:impact results}, BDB essentially expands the activation area, \emph{i.e.} uses as much spatial information as possible, to improve the performance. However, its random  %indiscriminately 
	feature dropping scheme makes all regions enhanced, regardless whether they are related to the target. This introduces the redundant features from irrelevant region.%, and even the misleading signals.
	
	This paper proposes a %Progressive Attentive feature Hard-Mix (PAHM) 
	Progressive Multi-stage feature Mix network (PMM), which consists of several branches ending with loss computation, %used as loss heads, 
	and intends to mine different local regions in each head to form the final representation. Formally, we design three sequential heads, with their inputs interacting with each other. The first one is provided with the original features (output of the backbone), while the second and third one are given the different features processed under the guidance of its previous stage. The key idea is to progressively suppress the over highlighted spatial regions in the input feature, and then force the next stage to discover other patterns different from those have already been exploited. To obtain the most salient spatial regions in the current stage, we employ the Grad-CAM \cite{gradcam} to back-propagate the correct logit score to its input feature map. Then the attentive feature Hard-Mix is applied to replace the highlighted blocks with those %the feature blocks 
	from the negative samples, thus the newly created feature can be a harder one, which has shown its significant impact in representation learning \cite{kalantidis2020hard}, and classification tasks \cite{softmargin_triplet, shrivastava2016training}. Experiments on the three %generally used 
	person reID benchmarks prove the effectiveness of the proposed methods.

	Our contributions are summarized as follows:
    \begin{itemize}
	\item We design a progressive multi-stage structure to suppress the most salient features for the current classifier, and force the head in the later stage to find other clues.
	\item We propose an attentive Hard-Mix feature augmentation method, which synthesizes the harder samples with mixing the negative pairs.
	\item We do intensive experiments on three different reID benchmarks, showing the effectiveness of our method.
    \end{itemize}
    
%-------------------------------------------------------------------------------
\section{Proposed Method}
	\label{proposition}
	%The proposed \emph{Progressive Attentive feaure Hard-Mix network} (PAHM)
	%This section includes three parts. 
	This section first introduces the progressive multi-stage structure. Then, we give a detailed description about the attentive Hard-Mix operation. Finally, a comparison between the proposed attentive Hard-Mix (A-Hard-Mix) and other augmentation methods, including BDB \cite{BFE}, CutMix \cite{CutMix} and attentive CutMix (A-CutMix) \cite{walawalkar2020attentive}, is provided.
%-------------------------------------------------------------------------------
	\subsection{Progressive Multi-Branch Structure}
	\label{framework}
    	\begin{figure*}[ht]
    		\centering
    		\includegraphics[width=0.8\textwidth]{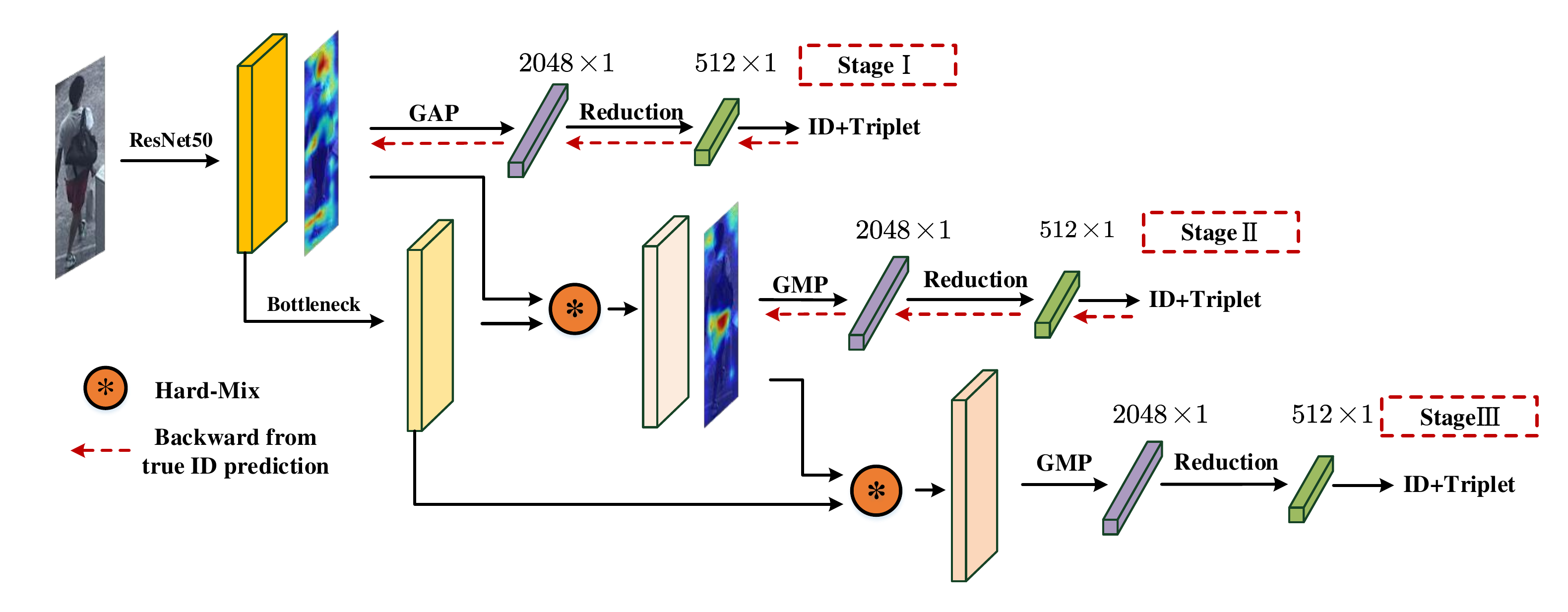}
    		\caption{Overview of the proposed Progressive Multi-stage feature Mix (PMM) Network. Three %independent 
    		stages are appended after the modified ResNet50 backbone, and they are supervised by both ID loss and triplet loss, respectively. Each stage can attain their own Grad-CAM images in the training period, which can be then used to guide the feature Hard-Mix in the next stage. In the testing, the green features from different stages are concatenated together as the final representation. }
    		\label{fig:framework}   
    	\end{figure*}
    	
    	To %the end of mining more diverse salient clues
    	discover different salient local parts, we suppress and mix features in a progressive way. %Figure \ref{fig:framework} shows the proposed progressive multi-stage structure. 
    	As shown in Figure \ref{fig:framework}, we adopt the modified ResNet50 \cite{resnet} as the backbone. It changes the down-sample stride to 1 %between the \emph{stage 3} and \emph{stage 4}
    	in the 3rd and 4th blocks, so that more spatial information can be preserved for the following multi-stage classifier. The same as BDB \cite{BFE}, each stage consists of one pooling layer, one reduction block ($1\times 1$ Conv/ BN/ ReLu) and one linear classification layer. Note that, global average pooling (GAP) is only applied to stage \uppercase\expandafter{\romannumeral1}, and global max pooling (GMP) is employed to both stage \uppercase\expandafter{\romannumeral2} and \uppercase\expandafter{\romannumeral3} to strengthen the relative weak features. To make the fair comparison, %with BDB, %we control the consistency of the number of network parameters
    	we keep a similar structure setting with BDB and use the same amount of learnable parameters, \emph{i.e.} we divide the second branch in BDB into stage \uppercase\expandafter{\romannumeral2} and \uppercase\expandafter{\romannumeral3}, and each has a 512-d embedding feature. %making a 1024-d embedding into two 512-d embeddings. 
    	
    	During the training phase, all the three stages are supervised by ID loss and the triplet loss \cite{softmargin_triplet}, %as is given in section \ref{sec:ls_f}, 
    	but fed with different input feature maps. For stage \uppercase\expandafter{\romannumeral1}, the input is directly taken from the backbone, while for stage \uppercase\expandafter{\romannumeral2} and \uppercase\expandafter{\romannumeral3}, the input is the mixed feature maps from two images based on the Grad-CAM image from the previous stage. As grad-CAM can locate the precise discriminative regions for the current stage, %suppressing and disturbing 
    	replacing these areas %stage by stage 
    	enforces the model to find other reliable features in the next stage.  %sequentially and in more detail. 
    	Note that we train the PMM network in an end-to-end manner, and the Grad-CAM is computed in the training loop. Since the stage  \uppercase\expandafter{\romannumeral2} and \uppercase\expandafter{\romannumeral3} require the guidance of Grad-CAM, twice back-propagations are conducted sequentially during the training. We also detach the gradients on Grad-CAM image, thus only the previous stage can influence the latter, while the latter stage is not able to distract the previous one. %to extract the most salient features it concerns. 
    	
    	In the testing phase, the attentive Hard-Mix operation is discarded as we cannot attain the correct guidance (gradients backward from the true prediction) at this period, and all the embedding features from different stages are concatenated together as the final descriptor.
%------------------------------------------------------------------------------

	\subsection{Attentive Hard-Mix}

	    \begin{figure}[ht]
    		\centering
    		\includegraphics[width=0.5\textwidth]{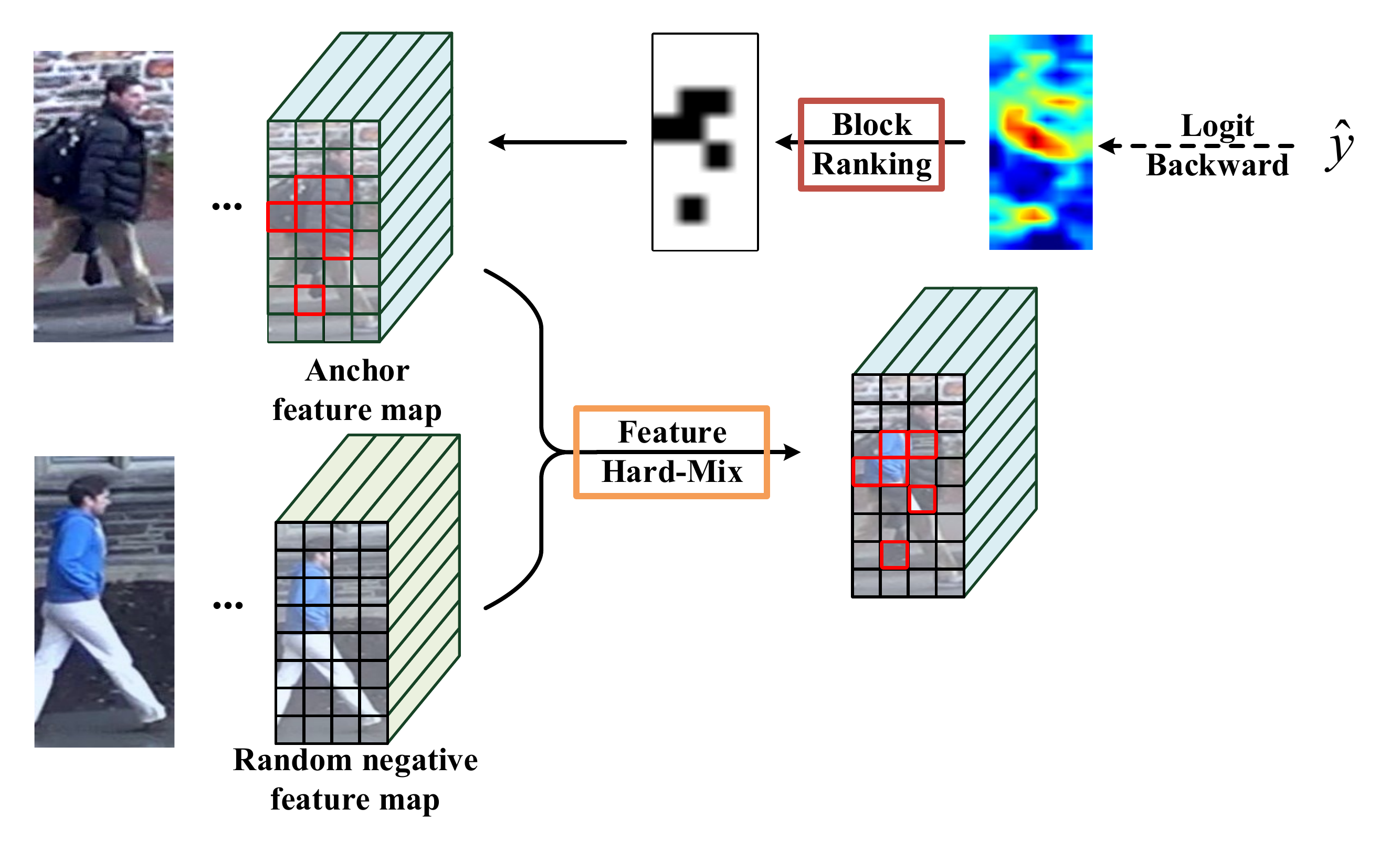}
    		\caption{The framework overview of the proposed Attentive feature Hard-Mix. %We divide it into 
    		There are two steps, \emph{i.e.} the block ranking step and the feature Hard-Mix step. Firstly, taking a Grad-CAM image, corresponding to the expected class, the block ranking %operation
    		turns it into a binary mask ( 0 for the highlighted regions in Grad-CAM and 1 for the rest). Then, according to the binary mask, the highlighted regions of an anchor feature map will be replaced by the corresponding region features from a random negative feature map in the feature Hard-Mix step.}
    		\label{fig:cutmix}
    	\end{figure}
    	
    	According to the visualization of %our observation
    	%the Grad-CAM image from 
    	BDB (see in Figure \ref{fig:impact results}), the heatmap is indeed enlarged, nevertheless, some irrelevant regions are also highlighted. %more unnecessary messages are also introduced. 
    	In order to solve this problem, we propose the Attentive feature Hard-Mix (A-Hard-Mix) to combine features from different classes based on %the guidance of 
    	the Grad-CAM image. %Training with 
    	The newly created samples are potentially hard for the current stage and they prevent the model making the prediction relying on the most dicriminative feature. %it enforce the network to pay more attention on the most discriminative areas. 
    	Figure \ref{fig:cutmix} illustrates the procedure of the A-Hard-Mix. It includes two steps. First, The Grad-CAM image $\bm{g}\in\mathcal{R}^{C\times H\times W}$, generated from the previous stage, is turned into a binary mask $\bm{m}\in\{0, 1\}^{H\times W}$ through the block ranking operation. Then, according to $\bm{m}$, a given anchor feature map $\bm{f}_a\in\mathcal{R}^{C\times H\times W}$ is to be cut and mixed with $\bm{f}_n\in\mathcal{R}^{C\times H\times W}$ from another image with different label. We now illustrate the block ranking and the Hard-Mix operations in the following two parts.
    	
        \textbf{Block Ranking.} A Grad-CAM image $\bm{g}$, computed from the gradients back-propagated from the expected class $\hat{y}$, is first spilt into the several blocks with the fixed size. We denote the blocks as $\bm{b}_i\in \mathbb{R}^{b_h\times b_w}$, where $i$ is the index of the block, $b_h$ and $b_w$ are the height and width of all the blocks. We specify that the width $W$ and height $H$ of $\bm{g}$ should be an integer multiple of $b_w$ and $b_h$, respectively. And there is no overlap between two adjacent blocks. Then, $\bm{b}_i$ are summed into a scalar $e_i$ to provide the ranking evidence of the $i$th block. Obviously, the higher value of $e_i$ indicates that the block $\bm{b}_i$ contributes a lot in the current classification stage, and following stages are intentionally designed for finding other discriminative regions. According to the ranking result, the top $K$ blocks in $\bm{m}$ are to be zeroed out, while the others are set as one, as shown in equation \ref{block ranking}. $j \in [1, H\times W]$ is the index on the spatial dimension of the generated mask.
    	
    	\begin{equation}
    	\label{block ranking}
        \bm{m}_j=\left\{
        \begin{array}{rcl}
        0 & & {j \in \{b_i\}\ \&\ e_i \in \{top_k(e)\}{}}\\
        1 & & others
        \end{array} \right.
        \end{equation}

    	\textbf{Feature Hard-Mix.} 
    	After acquiring the binary mask $\bm{m}$, we select the anchor feature $\bm{f_a}$ and the random negative $\bm{f_n}$ from different classes to make a mixture $\Tilde{\bm{f}}$. The mixing operation is formulated in equation \ref{cutmix}:
        	\begin{equation}
        	\label{cutmix}
        	\Tilde{\bm{f}} =  \bm{m}\odot\bm{f}_a + (1-\bm{m})\odot\bm{f}_n
    	\end{equation}	
    	Here $\odot$ indicates the spatial multiplication between the single-channel mask $\bm{m}$ and the $C$-channel feature map. The result $\Tilde{\bm{f}}$ is used as the input for the latter stage.

%------------------------------------------------------------------------------	  	
	\subsection{Comparison with Other Augmentation Methods}
	\label{differences with other augmentation methods}
	    \begin{table}[ht]
    	    \caption{Specific differences between the attentive Hard-Mix and other augmentation methods.}
    		\scalebox{0.8}{
            \begin{tabular}{l|c|c|c|c}
            \hline\hline
                                   & BDB     & CutMix & A-CutMix & A-Hard-Mix \\ \hline
            aug on feature / image & feature & image  & image    & feature    \\ %\hline
            attentive       & $\bigtimes$ &$\bigtimes$  & $\checkmark$ & $\checkmark$           \\ %\hline
            mix feature / image    & $\bigtimes$  & $\checkmark$ & $\checkmark$ & $\checkmark$           \\ %\hline
            mix label              & $\bigtimes$ & $\checkmark$ & $\checkmark$ & $\bigtimes$  \\          \hline \hline
            \end{tabular}}
            \label{tab:diff on different augmentation methods}
        \end{table}
	    
	    \textbf{Differences with BDB.} 
	    BDB \cite{BFE} takes a random drop feature augmentation to force the second stage to mine more diverse clues. Experiment results show that, BDB expands the high response area but the non-target region also contributes for the identification, which introduces redundant or even misleading information. The proposed attentive Hard-Mix is firstly guided by the Grad-CAM image, which provides the precise information of the current attentive area. Suppressing the current most salient features ensure that the features that have been fully utilized will not be intensively explored. On the contrary, those relatively weak features will be strengthened in the next stage. Secondly, the suppressed features specified by the Grad-CAM are replaced by the negative features from the same location. The attentive mixture of features from a negative pair is actually generating a harder example, which can be quite useful for representation learning \cite{kalantidis2020hard}.
	    %\newline
	    
	    \textbf{Differences with CutMix and Its Variant.} 
	    CutMix \cite{CutMix} is a data augmentation method used in the image-level. It proposes to take a random cut on a certain image, and paste the cut patch on a different image, where the ground-truth label is also mixed proportionally to the area of the patch. Attentive CutMix (A-CutMix) \cite{walawalkar2020attentive} is a variant of CutMix \cite{CutMix}, which takes the heatmap generated by a pretrained CNN to cut the most salient image patch, and then paste it to a different image. %The aim of this two methods 
	    Both of them aim %is both 
	    %to make a model 
	    to fully discover %the most 
	    salient and target-related features to make prediction. 
	    
	    %This 
	    Our idea %seems to be quite similar with ours, but there exist some
	    has significant differences with them. Firstly, the two methods %aim 
	    tend to strengthen the already well-learned regions, while our proposition %aims to 
	    suppresses the strong parts and focus on the relative weak region. Secondly, the two CutMix methods mix the labels, while A-Hard-Mix do not. In our setting, the most salient blocks %regions 
	    are removed and the replacement is not necessarily the %most 
	    salient part of the negative sample either, thus %the 
	    mixing %mixture on 
	    label may hinder the convergence, and introduce unnecessary noise. The detailed differences between the proposed A-Hard-Mix and the mentioned three augmentation are listed in Table \ref{tab:diff on different augmentation methods}. More results are provided in section \ref{sec:ablation study}.
%------------------------------------------------------------------------------
    \iffalse
	\subsection{Loss Functions.}\label{sec:ls_f}
	    We adopt the most widely used ID loss and triplet loss \cite{BFE, PCB, CAMA} to supervise the training process. The loss formulations are listed as below:
	    
	    \begin{equation}
	    \label{eq: id loss}
	    L_{id} = \sum_{i=1}^{N}-q_i\log(p_i)
	    \begin{cases}
        q_i=\epsilon / N & y\neq i\\
        q_i=1-\epsilon\frac{N-1}{N} & y=i
        \end{cases}
	    \end{equation}
	    
	    \begin{equation}
	    \label{eq: triplet loss}
	    L_{trip} = \sum_{i=1}^{N}[d_p-d_n+m]_+
	    \end{equation}
	    
	    \begin{equation}
	    \label{eq: total loss}
	    L = L_{id} + L_{trip}
	    \end{equation}
	    
	    ID loss is the cross entropy between the score $p_i$ after the softmax operation, and the smoothed groundtruth label $q_i$, which is proposed in . $\epsilon$ equals 0.1 to smooth the label. Triplet loss \cite{softmargin_triplet} mines the hardest negative and positive pairs on the concatenated feature embedding, and aims to pull the positive pairs closer (make the euclidean distance $d_p$ smaller), with pushing the negative ones further (make $d_n$ larger). In the above equations, $m$ is the margin, and $N$ is the batchsize. We sum the ID loss and the triplet loss together, and train the PMM with the total loss in an end-to-end way.
	\fi
	    
\section{Experiments}
\label{sec:experiments}
    \subsection{Datasets and Implementation Details}
    \label{experi: datasets and implementation}
    	We evaluate our proposed methods on three generally used person reID benchmarks, including Market-1501 \cite{market1501}, DukeM-TMC-reID \cite{duke}, CUHK03-Detected \cite{cuhk03}. We also take the standard strategy of the above datasets for generating the training set, gallery and the query \cite{BFE, PCB, CAMA}. For evaluation, we choose the Euclidean distance as the similarity measurement between query and gallery, and compute the Cumulative Matching Characteristic (CMC) curve \cite{cmc}. The evaluation metrics we adopted are Rank-1 accuracy and Mean Average Precision (mAP) \cite{market1501}. Notice that, re-ranking \cite{reranking} is not applied in our experiments. For the training process, we follow most of the configuration of BDB \cite{BFE}, except for the batchsize is 64 in our setting. The number of the selected attentive patches K is set as 3, 6, 8 for Market-1501, DukeMTMC-reID and CUHK03, respectively. And the block size in the block ranking operation is fixed as $3\times2$.

    \subsection{Results and Analysis}
        \subsubsection{Comparison with State-of-the-Art}
        	The comparison results between our proposition and other state-of-the-art methods are listed in Table \ref{comparison with sota}. To be fair, we train BDB \cite{BFE} with batch size 64 which is a little different from \cite{BFE} with batch size 128, and the results is denoted as \textbf{BDB*}. As shown in Table \ref{comparison with sota}, our method achieves great improvement compared with the previous work, especially for CUHK03 dataset, which is the most challenging one among the three. Under the same batch size setting, our method can get better performance with the BDB*. 
        	\begin{table}[ht]
        		\centering
        		\caption{Comparison results with the state-of-the-art methods on the classic reID datasets.}
        		\scalebox{0.8}{
        			\begin{tabular}{c|cc|cc|cc}
        				\hline\hline
        				{\multirow{2}{*}{Methods}} &  \multicolumn{2}{c|}{CUHK03} & \multicolumn{2}{c|}{DukeMTMC} & \multicolumn{2}{c}{Market} \\ \cline{2-7} 
        				& Rank-1           & mAP             & Rank-1           & mAP             & Rank-1         & mAP            \\ \hline
        				AOS \cite{AOS}         & 54.6            & 56.1           & 79.2            & 62.1           & 91.3          & 78.3          \\
        				PCB+RPP \cite{PCB}      & 62.8             & 56.7            & 83.3             & 69.2            & 93.8           & 81.6           \\
        				CAMA \cite{CAMA}  & 66.6             & 64.2            & 85.8    & 72.9   & \textbf{94.7}  & 84.5  \\  \hline
        				
        				BDB*\cite{BFE}       & 73.5                  & 69.8                & 87.1             & 74.5            & 94.0           & 84.9           \\
        				
        				Ours         & \textbf{76.3}                 & \textbf{73.0}                & \textbf{88.0}            & \textbf{74.6}            & 94.1           & \textbf{85.2}           \\ \hline\hline
        		\end{tabular}}		\label{comparison with sota}
        	\end{table}
	
%--------------------------------------------------------------------------
	\subsubsection{Ablation Studies}
	    \label{sec:ablation study}
    	%In this section, extensive experiments are conducted on Market-1501 and CUHK03-Detected to analyse the effectiveness of the proposed components.
%--------------------------------------------------------------------------

        \noindent\textbf{Effectiveness of the Attentive Hard-Mix.}
        
            Table \ref{tab: ablation on hardmix} shows the effectiveness of the proposed Attentive Hard-Mix. Compared with CutOut \cite{cutout}, we introduce the negative features and create new harder samples, which is effective for representation learning. As for CutMix and its variant, it can be hard to use the limited features to make precise prediction, thus the mixed label can bring some misleading information in the training process.
        
        % Please add the following required packages to your document preamble:
        % \usepackage{multirow}
        
            \begin{table}[ht]
            \centering
        		\caption{Comparison results between different feature augmentation methods.}
        		\scalebox{0.8}{
                \begin{tabular}{c|cc|cc|cc}
                \hline\hline
                    Methods    & \multicolumn{2}{l|}{Market-1501} & \multicolumn{2}{l|}{DukeMTMC} & \multicolumn{2}{l}{CUHK03} \\ \hline
                           & Rank-1          & mAP            & Rank-1          & mAP         & Rank-1        & mAP        \\ \hline
                CutOut     & 94.1            & 84.7           & 87.1           & 74.5         & 73.5          & 69.8       \\
                CutMix     & 92.8            & 82.4           & 85.8    & 72.0    &  74.3      & 69.9   \\
                A-CutMix   & 91.6            & 78.1           & 83.8   & 68.7  & 65.6          & 62.8       \\
                A-Hard-Mix & \textbf{94.1}   & \textbf{85.2}  & \textbf{88.0}       &\textbf{ 74.6 }      & \textbf{76.3}          & \textbf{73.0}       \\ \hline\hline
                \end{tabular}}
                \label{tab: ablation on hardmix}
            \end{table}
            
%--------------------------------------------------------------------------	
        %\newline
    	\noindent\textbf{Effectiveness of Each Stage.}
    	
\begin{figure}[ht]
        		\centering
        		\includegraphics[width=0.45\textwidth]{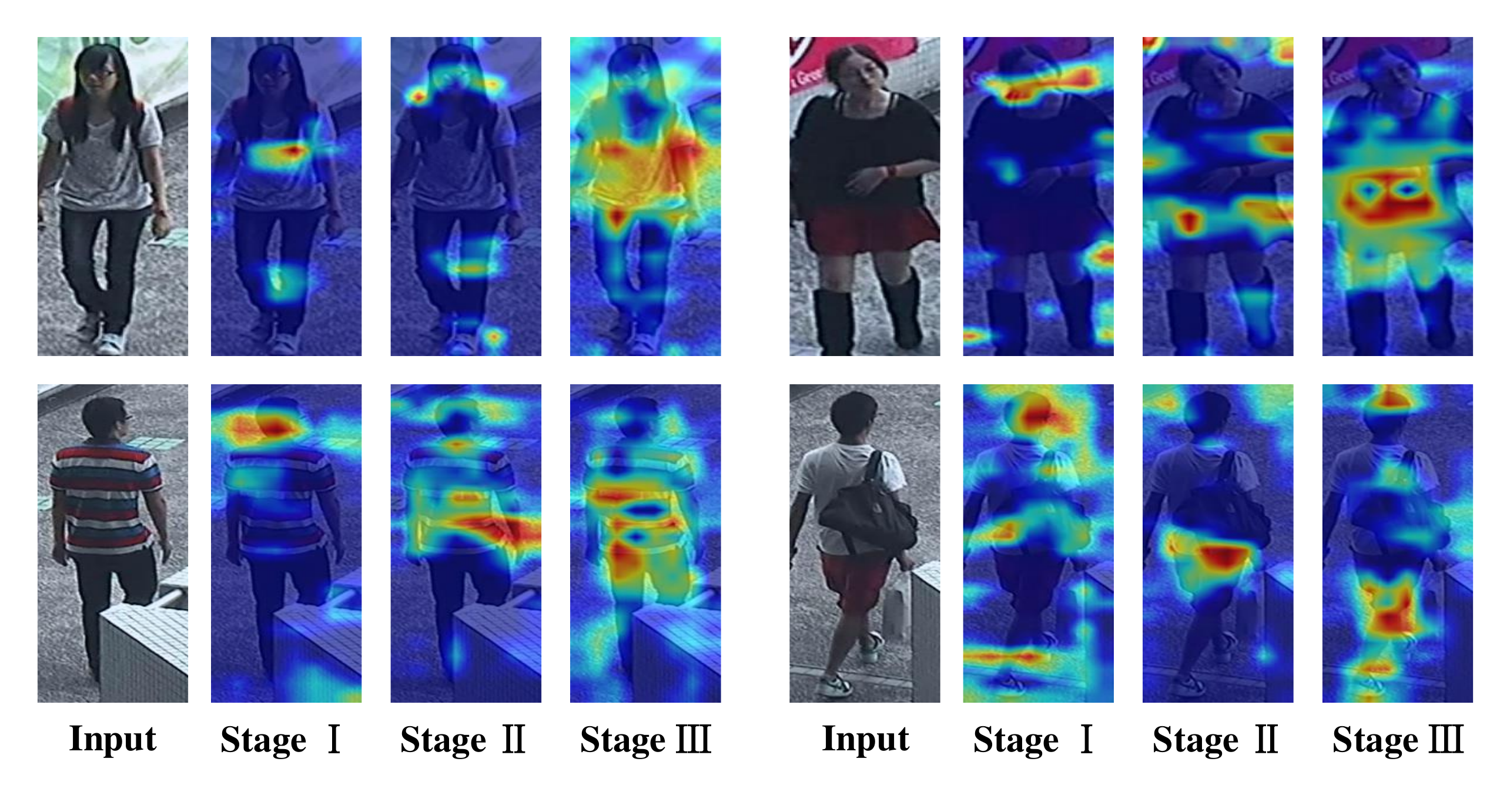}
        		\caption{The Grad-CAM image $\bm{g}$ from each stage %in our model 
        		during test (gradients are back-propagated from the predicted class). The visualization is conducted on CUHK03-Detected.} 
        		\label{fig:grad-CAM of each branch}
        	\end{figure}

        	The visualizations of the Grad-CAM of the three stages are shown in Figure \ref{fig:grad-CAM of each branch}. The activation maps learned by the three stages are different from each other, thus they can mutually provide complementary information. It is evident that, by suppressing and disturbing the highlighted regions, the response of features from the rest parts can be reinforced.
        	
%--------------------------------------------------------------------------

\section{Conclusion}
    In this paper, we propose a Progressive Multi-stage feature Mix network (PMM) for person re-identification. The %progressive multi-stage structure 
    PMM sequentially obstructs the most salient features stage by stage, enforcing the model to discover more diverse clues to form the final pedestrian representation. Meanwhile, the attentive Hard-Mix operation, which mixes the features from different classes together to create harder samples for training, significantly improves the performance. Experiment results meet our expectation and prove the effectiveness of our approach.

\newpage

% To start a new column (but not a new page) and help balance the last-page
% column length use \vfill\pagebreak.
% -------------------------------------------------------------------------
%\vfill
%\pagebreak

% References should be produced using the bibtex program from suitable
% BiBTeX files (here: strings, refs, manuals). The IEEEbib.bst bibliography
% style file from IEEE produces unsorted bibliography list.
% -------------------------------------------------------------------------
%\input{strings.bbl}
\bibliographystyle{IEEEbib}
\bibliography{strings}

\begin{thebibliography}{10}

\bibitem{BFE}
Zuozhuo Dai, Mingqiang Chen, Xiaodong Gu, Siyu Zhu, and Ping Tan,
\newblock ``Batch dropblock network for person re-identification and beyond,''
\newblock in {\em Proceedings of the IEEE International Conference on Computer
  Vision}, 2019, pp. 3691--3701.

\bibitem{topology}
Guan'an Wang, Shuo Yang, Huanyu Liu, Zhicheng Wang, Yang Yang, Shuliang Wang,
  Gang Yu, Jian Sun, et~al.,
\newblock ``High-order information matters: Learning relation and topology for
  occluded person re-identification,''
\newblock {\em arXiv preprint arXiv:2003.08177}, 2020.

\bibitem{su2017pose}
Chi Su, Jianing Li, Shiliang Zhang, Junliang Xing, Wen Gao, and Qi~Tian,
\newblock ``Pose-driven deep convolutional model for person
  re-identification,''
\newblock in {\em Proceedings of the IEEE international conference on computer
  vision}, 2017, pp. 3960--3969.

\bibitem{kumar2017pose}
Vijay Kumar, Anoop Namboodiri, Manohar Paluri, and CV~Jawahar,
\newblock ``Pose-aware person recognition,''
\newblock in {\em Proceedings of the IEEE Conference on Computer Vision and
  Pattern Recognition}, 2017, pp. 6223--6232.

\bibitem{zheng2019pose}
Liang Zheng, Yujia Huang, Huchuan Lu, and Yi~Yang,
\newblock ``Pose-invariant embedding for deep person re-identification,''
\newblock {\em IEEE Transactions on Image Processing}, vol. 28, no. 9, pp.
  4500--4509, 2019.

\bibitem{kalayeh2018human}
Mahdi~M Kalayeh, Emrah Basaran, Muhittin G{\"o}kmen, Mustafa~E Kamasak, and
  Mubarak Shah,
\newblock ``Human semantic parsing for person re-identification,''
\newblock in {\em Proceedings of the IEEE Conference on Computer Vision and
  Pattern Recognition}, 2018, pp. 1062--1071.

\bibitem{gradcam}
Ramprasaath~R. Selvaraju, Abhishek Das, Ramakrishna Vedantam, Michael Cogswell,
  Devi Parikh, and Dhruv Batra,
\newblock ``Grad-cam: Why did you say that? visual explanations from deep
  networks via gradient-based localization,''
\newblock {\em CoRR}, vol. abs/1610.02391, 2016.

\bibitem{kalantidis2020hard}
Yannis Kalantidis, Mert~Bulent Sariyildiz, Noe Pion, Philippe Weinzaepfel, and
  Diane Larlus,
\newblock ``Hard negative mixing for contrastive learning,''
\newblock {\em arXiv preprint arXiv:2010.01028}, 2020.

\bibitem{softmargin_triplet}
Alexander Hermans, Lucas Beyer, and Bastian Leibe,
\newblock ``In defense of the triplet loss for person re-identification,''
\newblock {\em arXiv preprint arXiv:1703.07737}, 2017.

\bibitem{shrivastava2016training}
Abhinav Shrivastava, Abhinav Gupta, and Ross Girshick,
\newblock ``Training region-based object detectors with online hard example
  mining,''
\newblock in {\em Proceedings of the IEEE conference on computer vision and
  pattern recognition}, 2016, pp. 761--769.

\bibitem{CutMix}
Sangdoo Yun, Dongyoon Han, Seong~Joon Oh, Sanghyuk Chun, Junsuk Choe, and
  Youngjoon Yoo,
\newblock ``Cutmix: Regularization strategy to train strong classifiers with
  localizable features,''
\newblock in {\em Proceedings of the IEEE International Conference on Computer
  Vision}, 2019, pp. 6023--6032.

\bibitem{walawalkar2020attentive}
Devesh Walawalkar, Zhiqiang Shen, Zechun Liu, and Marios Savvides,
\newblock ``Attentive cutmix: An enhanced data augmentation approach for deep
  learning based image classification,''
\newblock in {\em ICASSP 2020-2020 IEEE International Conference on Acoustics,
  Speech and Signal Processing (ICASSP)}. IEEE, 2020, pp. 3642--3646.

\bibitem{resnet}
Kaiming He, Xiangyu Zhang, Shaoqing Ren, and Jian Sun,
\newblock ``Deep residual learning for image recognition,''
\newblock in {\em Proceedings of the IEEE conference on computer vision and
  pattern recognition}, 2016, pp. 770--778.

\bibitem{market1501}
Liang Zheng, Liyue Shen, Lu~Tian, Shengjin Wang, Jingdong Wang, and Qi~Tian,
\newblock ``Scalable person re-identification: A benchmark,''
\newblock in {\em Computer Vision, IEEE International Conference on}, 2015.

\bibitem{duke}
Ergys Ristani, Francesco Solera, Roger Zou, Rita Cucchiara, and Carlo Tomasi,
\newblock ``Performance measures and a data set for multi-target, multi-camera
  tracking,''
\newblock in {\em European Conference on Computer Vision}. Springer, 2016, pp.
  17--35.

\bibitem{cuhk03}
Wei Li, Rui Zhao, Tong Xiao, and Xiaogang Wang,
\newblock ``Deepreid: Deep filter pairing neural network for person
  re-identification,''
\newblock in {\em Proceedings of the IEEE conference on computer vision and
  pattern recognition}, 2014, pp. 152--159.

\bibitem{PCB}
Yifan Sun, Liang Zheng, Yi~Yang, Qi~Tian, and Shengjin Wang,
\newblock ``Beyond part models: Person retrieval with refined part pooling (and
  a strong convolutional baseline),''
\newblock in {\em Proceedings of the European Conference on Computer Vision
  (ECCV)}, 2018, pp. 480--496.

\bibitem{CAMA}
Wenjie Yang, Houjing Huang, Zhang Zhang, Xiaotang Chen, Kaiqi Huang, and Shu
  Zhang,
\newblock ``Towards rich feature discovery with class activation maps
  augmentation for person re-identification,''
\newblock in {\em Proceedings of the IEEE Conference on Computer Vision and
  Pattern Recognition}, 2019, pp. 1389--1398.

\bibitem{cmc}
Douglas Gray, Shane Brennan, and Hai Tao,
\newblock ``Evaluating appearance models for recognition, reacquisition, and
  tracking,''
\newblock in {\em Proc. IEEE international workshop on performance evaluation
  for tracking and surveillance (PETS)}. Citeseer, 2007, vol.~3, pp. 1--7.

\bibitem{reranking}
Zhun Zhong, Liang Zheng, Donglin Cao, and Shaozi Li,
\newblock ``Re-ranking person re-identification with k-reciprocal encoding,''
\newblock in {\em CVPR}, 2017.

\bibitem{AOS}
Houjing Huang, Dangwei Li, Zhang Zhang, Xiaotang Chen, and Kaiqi Huang,
\newblock ``Adversarially occluded samples for person re-identification,''
\newblock in {\em Proceedings of the IEEE Conference on Computer Vision and
  Pattern Recognition}, 2018, pp. 5098--5107.

\bibitem{cutout}
Terrance DeVries and Graham~W Taylor,
\newblock ``Improved regularization of convolutional neural networks with
  cutout,''
\newblock {\em arXiv preprint arXiv:1708.04552}, 2017.

\end{thebibliography}

\end{document}